%% file: main.tex
\documentclass[conference]{IEEEtran}
\IEEEoverridecommandlockouts
\usepackage{cite}
\usepackage{amsmath,amssymb,amsfonts}
\usepackage{algorithmic}
\usepackage{graphicx}
\usepackage{textcomp, gensymb}
\usepackage{xcolor}
\usepackage[pagebackref=false,breaklinks=true,letterpaper=true,colorlinks=false,citecolor=black,bookmarks=false]{hyperref}

\def\BibTeX{{\rm B\kern-.05em{\sc i\kern-.025em b}\kern-.08em
    T\kern-.1667em\lower.7ex\hbox{E}\kern-.125emX}}

\usepackage[utf8]{inputenc}

\newcommand{\gqcnn}{GQ-CNN}
\newcommand{\ours}{VGQ-CNN} 
\newcommand{\fast}{Fast-\ours} 
\newcommand{\dset}{VG-dset} 
\newcommand{\ReducedDset}{TG-Tset} 
\include{nomenclature}

\begin{document}

\title{\ours: Moving Beyond Fixed Cameras and Top-Grasps for Grasp Quality Prediction\\
\thanks{This publication has emanated from research supported in part by Grants from Science Foundation Ireland under Grant numbers 18/CRT/6049 and 16/RI/3399.}
}

\author{\IEEEauthorblockN{Anna Konrad}
\IEEEauthorblockA{\textit{Hamilton Institute} \\
\textit{Maynooth University}\\
anna.konrad.2020@mumail.ie}
\and
\IEEEauthorblockN{John McDonald}
\IEEEauthorblockA{\textit{Department of Computer Science} \\
\textit{Maynooth University}\\
john.mcdonald@mu.ie}
\and
\IEEEauthorblockN{Rudi Villing}
\IEEEauthorblockA{\textit{Department of Electronic Engineering} \\
\textit{Maynooth University}\\
rudi.villing@mu.ie}
}

\maketitle

\begin{abstract}
We present the Versatile Grasp Quality Convolutional Neural Network (\ours), a grasp quality prediction network for {6-DOF} grasps. \ours\ can be used when evaluating grasps for objects seen from a wide range of camera poses or mobile robots without the need to retrain the network. By defining the grasp orientation explicitly as an input to the network, \ours\ can evaluate {6-DOF} grasp poses, moving beyond the {4-DOF} grasps used in most image-based grasp evaluation methods like \gqcnn. To train \ours, we generate the new Versatile Grasp dataset (\dset) containing {6-DOF} grasps observed from a wide range of camera poses. \ours\ achieves a balanced accuracy of 82.1$\%$ on our test-split while generalising to a variety of camera poses. Meanwhile, it achieves competitive performance for overhead cameras and top-grasps with a balanced accuracy of 74.2$\%$ compared to \gqcnn's 76.6$\%$. We also propose a modified network architecture, \fast, that speeds up inference using a shared encoder architecture and can make 128 grasp quality predictions in 12ms on a CPU. Code and data are available at \url{https://aucoroboticsmu.github.io/vgq-cnn/}.
\end{abstract}


\section{Introduction}

Robotic grasping is a persistent challenge within robotics research due to its inherent complexity. While robotic grasping techniques are widely used in industrial setups where the influence of uncertainties can be reduced, the usage in less structured environments is still an active field of research. Multiple different approaches for grasping unknown objects have been developed, with the camera usually fixed to a pose above the object and the robot in a stationary pose close to the camera~\cite{mahler2017dexnet2, levine2018, zhang2020randomforest, redmon2015, kumra2020, song2020graspdetection}. These fixed relationships between camera, object and robot limit the applicability of the trained networks in non-lab scenarios. When attempting to apply such networks to a robot with a different camera pose or when working with multiple cameras, the networks have to be retrained each time as they do not generalise to versatile viewpoints. Furthermore, usage with mobile manipulators (e.g. the PAL TIAGo, Toyota HSR, etc.), which constantly change their pose in relation to the object, is not practical if constrained to fixed camera poses.

Another related limitation is the use of {4-DOF} grasp representations in current image-based approaches. By assuming alignment between the grasps and the table normal or the camera principal axis~\cite{mahler2017dexnet2, zhang2020randomforest, satish2019policy, morrison2019, redmon2015, kumra2020, song2020graspdetection}, the grasps can be represented as a {2-D} position and {1-D} orientation within the image frame. In commonly reported configurations that mount the camera above the object~\cite{mahler2017dexnet2, satish2019policy, depierre2018jacquard, kumra2020} these networks are limited to top-grasps, where the grasp approach is along the table normal. Consequently, the variety of grasp orientations is limited significantly, creating constraints for possible path planning pipelines downstream.

We base our work on \gqcnn\ by Mahler et al.~\cite{mahler2017dexnet2}, a neural network for grasp quality evaluation of top-grasps from an overhead camera. While there have been follow-up networks for \gqcnn~\cite{mahler2017clutter, mahler2017dexnet3, mahler2019dexnet4}, these extensions have focused on objects in clutter and alternative gripper mechanisms. As such, these networks maintain the {4-DOF} grasp representation and therefore are also limited to top-grasps. While a fully convolutional version of \gqcnn\ improves run-time~\cite{satish2019policy}, the necessary parameterisation to allow for {6-DOF} grasps would lead to a significant increase in the size of the output tensor as currently formulated.

\begin{figure*}[t]
      \centering
      \includegraphics[width=0.55\textwidth]{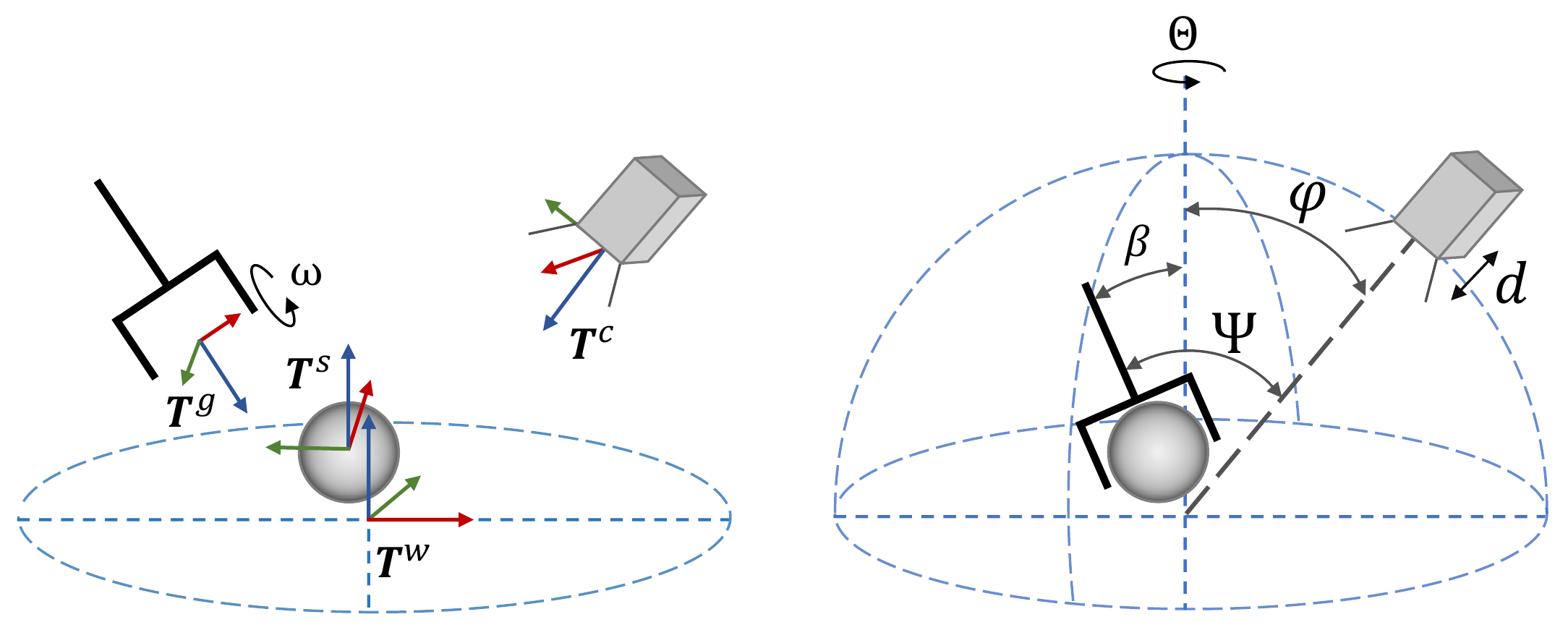}
      \caption{Visualisation of coordinate frames, parameters and relative angles of camera-object-gripper configurations.}
      \label{fig:sphere}
\end{figure*}

We move \gqcnn\ beyond fixed cameras and top-grasps with the Versatile Grasp Quality Convolutional Neural Network (\ours). The network can evaluate 6-DOF grasp proposals of objects on a planar surface viewed from a wide range of camera poses above the object. The position and viewing angle of the camera can be varied within a space as large as $2.1m^3$. We introduce the Versatile Grasp dataset (\dset) to include these variations in camera poses and grasp orientations, significantly exceeding the range available in commonly used datasets such as~\cite{mahler2017dexnet2, depierre2018jacquard, lenz2015}. We further propose using a shared image encoder with our alternative network architecture, \fast, to speed up the evaluation of multiple grasps during run-time. In summary, the contributions of this paper are:
\begin{itemize}
    \item \dset: A versatile {6-DOF} grasp quality dataset including $7.1$ million grasps over an extended range of camera poses.
    \item \ours: A network capable of predicting the quality of 6-DOF grasps under a wide range of camera poses.
    \item \fast: An alternative network structure to \ours\ to speed up inference and predict up to 128 grasps in $8ms$ on a NVIDIA RTX 2060 GPU and $12ms$ on an Intel i7-10750H CPU.
\end{itemize}

\section{Related work}\label{sec:background}

Identifying good grasping poses for unknown objects has been an active field of research for many years~\cite{bohg2014, kleeberger2020survey}. Grasps can either be directly generated via generative methods~\cite{morrison2019, satish2019policy, breyer2020volumetric, kumra2020} or sampled and subsequently ranked by discriminative methods like \gqcnn~\cite{mahler2017dexnet2, levine2018, pas2017pointclouds, mousavian20196dof}. Discriminative approaches are typically coupled with a higher runtime and more flexibility, while the parameterisation in generative approaches allows for low latency while limiting flexibility~\cite{kleeberger2020survey}. Independent of the approach, many of the presented methods have been developed for a camera pose fixed within a small tolerance in relation to the object~\cite{mahler2017dexnet2, levine2018, kumra2020, satish2019policy}.

In addition to restricted camera poses, the methods usually only allow for {4-DOF} grasps. This is especially apparent when taking into account the grasp representations utilised in those methods, often projecting the Tool Centre Point (TCP) and gripper orientation into the image plane and defining it as an oriented rectangle~\cite{kumra2020, depierre2018jacquard, lenz2015, morrison2019} or pre-processing the image to align it with the grasp~\cite{mahler2017dexnet2, satish2019policy}. Due to this representation the variety of grasps is reduced significantly, often only allowing for top-grasps~\cite{mahler2017dexnet2, satish2019policy, kumra2020, depierre2018jacquard}.

When moving away from those grasp representations and towards unrestricted {6-DOF} grasps, the dimensions and complexity of the problem increase. Methods providing this functionality usually utilise point clouds and process them directly~\cite{mousavian20196dof}, in a voxelised grid~\cite{pas2017pointclouds, breyer2020volumetric} or as object meshes through shape completion~\cite{lundell2020scenecompletion, varley2017shapecompletion}. One of the major disadvantages that working with unrestricted {6-DOF} grasps often poses is increased run-time, typically taking several seconds~\cite{mousavian20196dof, pas2017pointclouds, lundell2020scenecompletion}. While generative methods for {6-DOF} grasps can reduce run-time down to $10ms$ with specialised hardware~\cite{breyer2020volumetric, berscheid20216dof}, the parameterisation limits the number and flexibility of proposed grasps.

Another way {6-DOF} grasps can be facilitated is by attaching cameras to the wrist of robotic manipulators and then iteratively approaching the object~\cite{viereck2017, song2020grasping}. 
These approaches might be difficult to incorporate into the overall path planning and collision checking of a robot, since they produce control outputs~\cite{viereck2017} or actions~\cite{song2020grasping} rather than goal poses. Instead, they could be used for a closed-loop approach of the grasp pose after a pre-grasp position provided by an overall grasp planner has been reached.

Robotic grasping for real-world applications has to be fast and flexible in order to be used efficiently. Approaches need versatility in the configuration of grasp orientations and camera poses to cater for dynamic setups. Recent approaches have developed generative {6-DOF} grasp proposal methods based on depth images~\cite{zhu2021graspproposal}. While such approaches can compute grasp proposals within $0.5s$, they employ a camera positioned directly above the object and hence are not applicable to situations involving a wider variation of camera poses, e.g. with mobile manipulators. Extending such work to a discriminative grasp sampling method can enhance flexibility in the number of grasps that can be proposed. This enables not only iterative refinement of grasps~\cite{mousavian20196dof}, but could also be used to tailor grasp sampling on specific objects or object areas. Following this approach, we present \ours, a {6-DOF} grasp quality prediction network based on depth images that is robust to a wide range of camera poses. We furthermore provide our alternative network architecture \fast, which speeds up grasp prediction significantly to enhance usability for real-world applications.

\section{Problem formulation}\label{sec:problem_formulation}

We consider the problem of predicting grasp qualities for flexible grasp orientations with a parallel jaw gripper as observed from a wide range of camera perspectives. The environment is limited to a single object placed on a planar surface. The goal is to train a network which can predict grasp qualities based on depth images and grasp proposals from a wide range of camera poses without having to retrain the network for each new camera pose. In addition, the network should be able to evaluate a variety of {6-DOF} grasp orientations where the gripper approach axis is not necessarily aligned with a pre-specified fixed axis, e.g. the camera principal ray. Such a network caters for scenarios that involve changing the pose of the object or camera between setups, whilst also including mobile manipulators, where the relationship between camera and object varies dynamically.

A visualisation of the setup can be seen in Fig.~\ref{fig:sphere}. The object meshes are placed on a planar surface in predefined, stable resting poses, with their coordinate frame $T^s$ being in close proximity to the origin of world coordinate frame $T^w$. The camera position is defined in spherical coordinates in relation to $T^w$, with the distance to the origin $\cameraradius$, the elevation angle $\elev$ and the polar angle $\polar$ defining its position. The camera principal ray is pointing towards the origin of $T^w$ and the x-axis is parallel to the table surface, orienting the camera frame, $T^c$, such that the camera views the table and object horizontal and upright, respectively. The gripper frame, $T^g$, is defined such that the x-axis lies between the contact points of the parallel jaw gripper and the z-axis denotes the linear approach direction.

We define the camera as placed above the planar surface with $\cameraradius \in [0.4m, 1.1m]$, $\polar \in [0\degree, 360\degree]$ and $\elev \in [0\degree, 70\degree]$, which corresponds to a volume of roughly $2.1m^3$. The limits were chosen in reference to the typical accessible range of mobile manipulators, e.g. {PAL} TIAGo robot~\cite{pal2016tiago}, Toyota HSR~\cite{yamamoto2018toyotahsr}, grasping objects placed on a table. Configurations where the camera is below or level with the planar surface, e.g. when grasping objects from a shelf, and those with objects placed close to the edge of the surface are excluded.

Since the quality and usability of a grasp is influenced not only by the gripper pose in relation to the table and the object, but also by the visibility from the camera, we define a set of relative angles to parameterise the space of unique camera-object-gripper configurations. The angle between the gripper z-axis and the table normal is denoted as $\grasptableangle$, while the angle between the gripper z-axis and the camera principal ray is denoted as $\graspcameraangle$. Rotating the grasp around the grasp x-axis is denoted by $\grasphinge$ and does not alter the position of the contact points. For viable {6-DOF} grasp poses, the angle between the gripper z-axis and the table normal is constrained to $\grasptableangle \in [0\degree, 90\degree]$. Since $\grasptableangle \geq 90\degree$ would likely cause collisions with the table, we exclude such grasps from our setup. 

\begin{table}[tb]
\centering
\caption{Parameter overview}\label{tab:dset}
\vspace{-0.3cm}
\begin{tabular}{c|c c c c}
Dataset & $\elev_{max}$  & $\cameraradius_{min}$ &  $\cameraradius_{max}$ & $\grasptableangle_{max}$\\
\hline
\dset & $70\degree$ & $0.4m$ & $1.1m$ & $90\degree$\\
DexNet2.0 & $5.7\degree$ & $0.65m$ & $0.75m$ & $5\degree$\\
\end{tabular}
\end{table}

\section{Dataset}\label{sec:dataset}

We create a new 6-DOF grasp quality dataset called the Versatile Grasp dataset (\dset) to satisfy the specifications in our problem formulation in section~\ref{sec:problem_formulation}. The complete overview of parameter ranges for \dset\ in comparison to DexNet2.0~\cite{mahler2017dexnet2} can be seen in Table~\ref{tab:dset}. We base \dset\ on the pre-sampled antipodal grasps $g \in \mathcal{G}(o)$ included in DexNet2.0~\cite{mahler2017dexnet2}, where the object meshes $o \in \mathcal{O}$ are taken from the KIT~\cite{kasper2012kit} and 3DNet~\cite{wohlkinger20123dnet} mesh datasets.

Dataset preparation consists of two stages: dataset rendering and dataset sampling. Throughout the dataset rendering process, datapoints are created by rendering images and storing grasp poses and grasp quality values. A total of 131 million grasps are stored in the process, providing a baseline which can be sampled for different purposes. The dataset sampling process is necessary since certain minimal ratios, e.g. between ground-truth negative and ground-truth positive grasps, have to be satisfied to ensure successful training of \ours. The process can also be used to allow for different dataset compositions, e.g. excluding certain camera or grasp configurations. While the dataset generation process could be adjusted to generate a single, balanced dataset to be used for training, the time-consuming process of rendering the data and checking the grasps would have to be repeated each time a different dataset composition was desired. 

\subsection{Dataset rendering}\label{sec:dset_rendering}

\begin{figure}[t]
      \centering
      \includegraphics[width=0.5\textwidth]{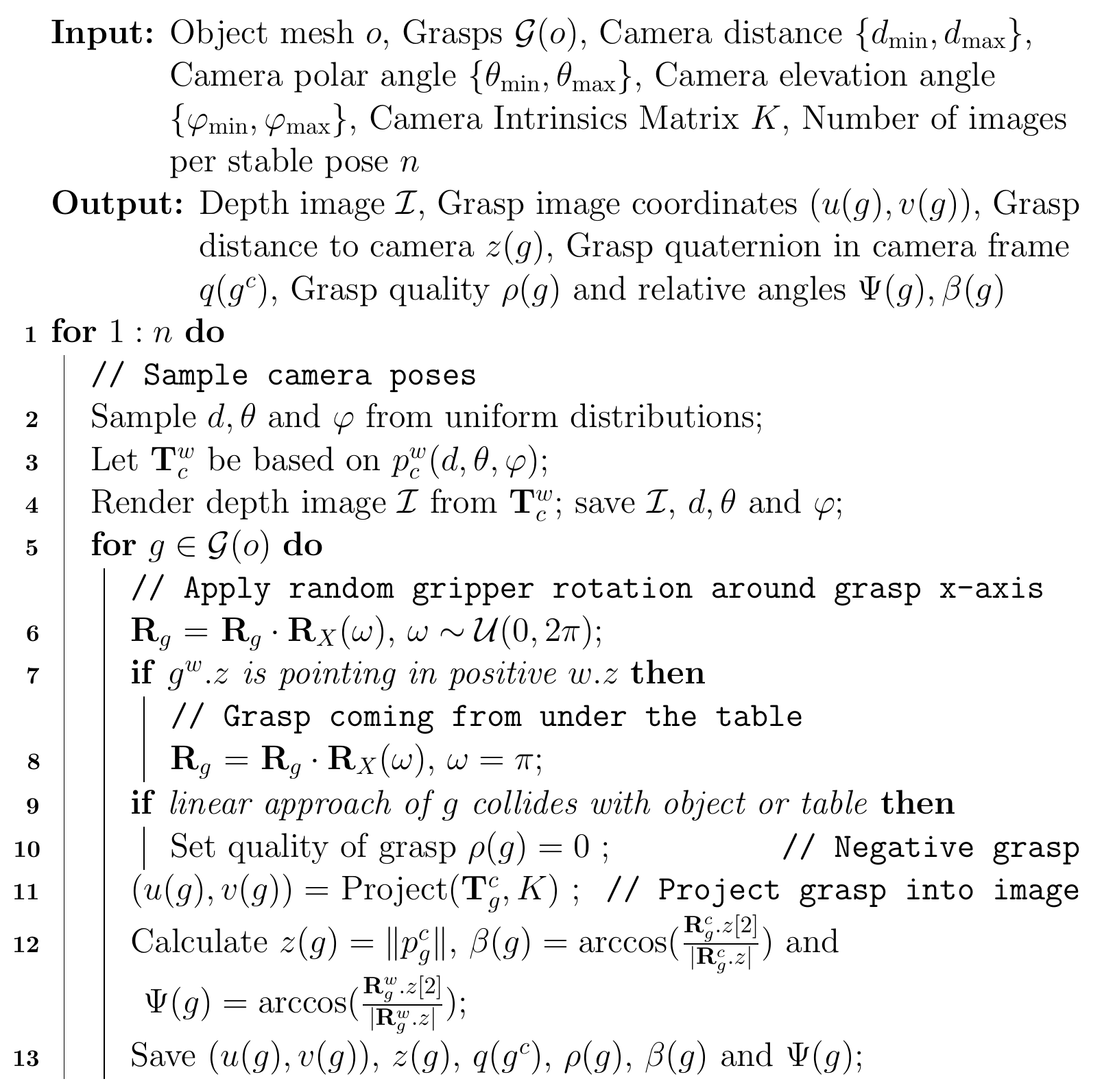}
      \caption{Algorithm for rendering \dset.}
      \label{fig:alg}
\end{figure}

We base our dataset rendering algorithm on the DexNet2.0 implementation by Mahler et. al.~\cite{mahler2017dexnet2} available on github\footnote{https://github.com/BerkeleyAutomation/dex-net} with key changes in the camera poses, grasp alignment and grasp representation.  Each object mesh $o \in \mathcal{O}$ has an average of $9$ stable resting poses $s \in \mathcal{S}(o)$. When rendering the dataset, we repeat the algorithm detailed in Fig.~\ref{fig:alg} for each stable resting pose of each of the $1494$ object meshes.

Moving away from top-grasps and allowing for a variety of grasp orientations represents a challenge to both the grasp preparation and representation. For stationary cameras and top-grasps, some sense of alignment between the grasp orientation and the camera position is usually assumed. In DexNet2.0~\cite{mahler2017dexnet2}, this is realised by rotating the gripper around $\grasphinge$ to minimise $\grasptableangle$ and discarding grasps with $\grasptableangle \geq 5 \degree$. By projecting the gripper Tool Centre Point (TCP) and gripper x-axis into the image plane and rotating/cropping the image accordingly, Mahler et al.~\cite{mahler2017dexnet2} reduce the grasp representation to an aligned image and the distance between the grasp $T^g$ and the camera $T^c$. This is similar to the reduced grasp representations in~\cite{zhang2020randomforest, redmon2015, morrison2019}, all of which are defined as positions and rotations in the {2-D} image plane.

Since \dset\ should include {6-DOF} grasps, we aim to include grasps with the same grasp x-axis and varying approach angles. For example, a top-grasp rotated by $90\degree$ around $\grasphinge$ would end up grasping the object parallel to the table. Ideally, both grasps should be included and proposed to the path planning system of a robot to choose the best grasp. We augment the data with a variety of grasp orientations in \dset\ by applying a random rotation around $\grasphinge$ to the grasps with each new camera pose. We flip grasps approaching from under the table with $\grasptableangle > 90\degree$ by rotating them once more for $180\degree$ around $\grasphinge$. We use this approach to provide flexibility to various sampling schemes (see section~\ref{sec:dset_sampling}) and to simplify constraining the final grasp orientation. Note that rotating the gripper around $\grasphinge$ does not change the robust Ferrari-Canny metric~\cite{seita2016grasp} of a given grasp, and therefore does not influence the grasp quality $\graspquality(g)$ aside from collisions checked after the final grasp orientation has been decided.

We then, like in DexNet2.0~\cite{mahler2017dexnet2}, proceed to collision checking (Fig.~\ref{fig:alg}, lines 9-10), setting the robust Ferrari-Canny value for grasps colliding with the object or table $\graspquality(g) = 0$ and thereby marking the grasp as ground-truth negative. After collision checking, the gripper TCP is projected into the image plane in order to calculate its $(u, v)$ position in the image in pixel coordinates. 

Note that we render $n = 100$ images for each stable resting pose $s \in \mathcal{S}(o)$ of each object mesh $o \in \mathcal{O}$, while $n = 50$ for the DexNet2.0 dataset generation. Since our data includes a wider range of camera poses and grasp orientations, sampling more images per stable pose allows us to subsample the resulting dataset in order to create specific dataset characteristics as explained in the following section.

\subsection{Dataset sampling}\label{sec:dset_sampling}

The 130.8 million grasps from the dataset rendering stage need to be undersampled to generate \dset, a dataset for \ours. This undersampling process is not needed in DexNet2.0~\cite{mahler2017dexnet2} due to their dataset parameters as described in Table~\ref{tab:dset}. To train a network on our new problem formulation, undersampling the datapoints becomes necessary for three major reasons:

\begin{itemize}
    \item Filtering grasp and/or camera configurations for training \ours, e.g. removing grasps with $\graspcameraangle \geq 90\degree$.
    \item Adjusting the positivity rate $pos_r = \frac{100 \times\#positive\ grasps}{\#all\ grasps}$, since it affects the training loss and therefore the network performance.
    \item Balancing the number of grasps across the camera and grasp configurations.
\end{itemize}

The large size of the rendered dataset enables differing dataset compositions to be sampled and their effect on the network performance to be tested.

To generate \dset, grasps with $\graspcameraangle \geq 90\degree$ are removed from the dataset. This maximum value is set so that grasps approaching the object from behind relative to the camera position are excluded, since they are likely to be occluded by the object. 
Second, we undersample negative grasps to ensure a consistent positivity rate $pos_r$ over beta. Prior to undersampling, $pos_r$ declines with increasing $\grasptableangle$, since grasps with a high $\grasptableangle$ tend to collide with the table more often. The native, rendered dataset has a varying positivity rate with a mean of $pos_r = 6\%$, while the positivity rate for DexNet2.0 is $pos_r = 19\%$. 
We adjust $pos_r$ by undersampling negative/positive grasps based on the sampling rate $sample\_rate_{x}$. Since $pos_r$ is dependent on $\grasptableangle$, we calculate the sampling rate for undersampling negative grasps as,
\begin{align}
sample\_rate_{neg}(\Delta\grasptableangle) = \frac{\frac{100}{19} \times pos_r(\Delta\grasptableangle) - pos_r(\Delta\grasptableangle)}{100 - pos_r(\Delta\grasptableangle)} \nonumber
\end{align}
where, $\Delta\grasptableangle$ is varied in steps of $5\degree$. The undersampling rate for positive grasps is set to $sample\_rate_{pos} = \frac{1}{sample\_rate_{neg}}$. Undersampling rates of $sample\_rate_{x} > 1$ are set to 1. We then skip positive/negative grasps randomly based on their sampling rate for the given $\grasptableangle$. 
As a third step, we undersample the remaining grasps to have a uniform sample size over $\elev$ and $\grasptableangle$. The resulting dataset, \dset, consists of 7.2 million grasps. As shown in section~\ref{sec:res_ablation}, training on datasets with 2 million or more grasps shows a constant performance, while training on dataset with fewer than 1 million grasps shows signs of overfitting the dataset on \ours.

We divide both DexNet2.0 and \dset\ in an object-wise training, validation and test split of 80-10-10, using the same objects for the test sets in DexNet2.0 and \dset. This allows us to compare performance between \ours\ and \gqcnn\ on the same test data in section~\ref{sec:res_overall}.

\section{Network architecture}\label{sec:architecture}

\begin{figure}[t]
      \centering
      \includegraphics[width=0.48\textwidth]{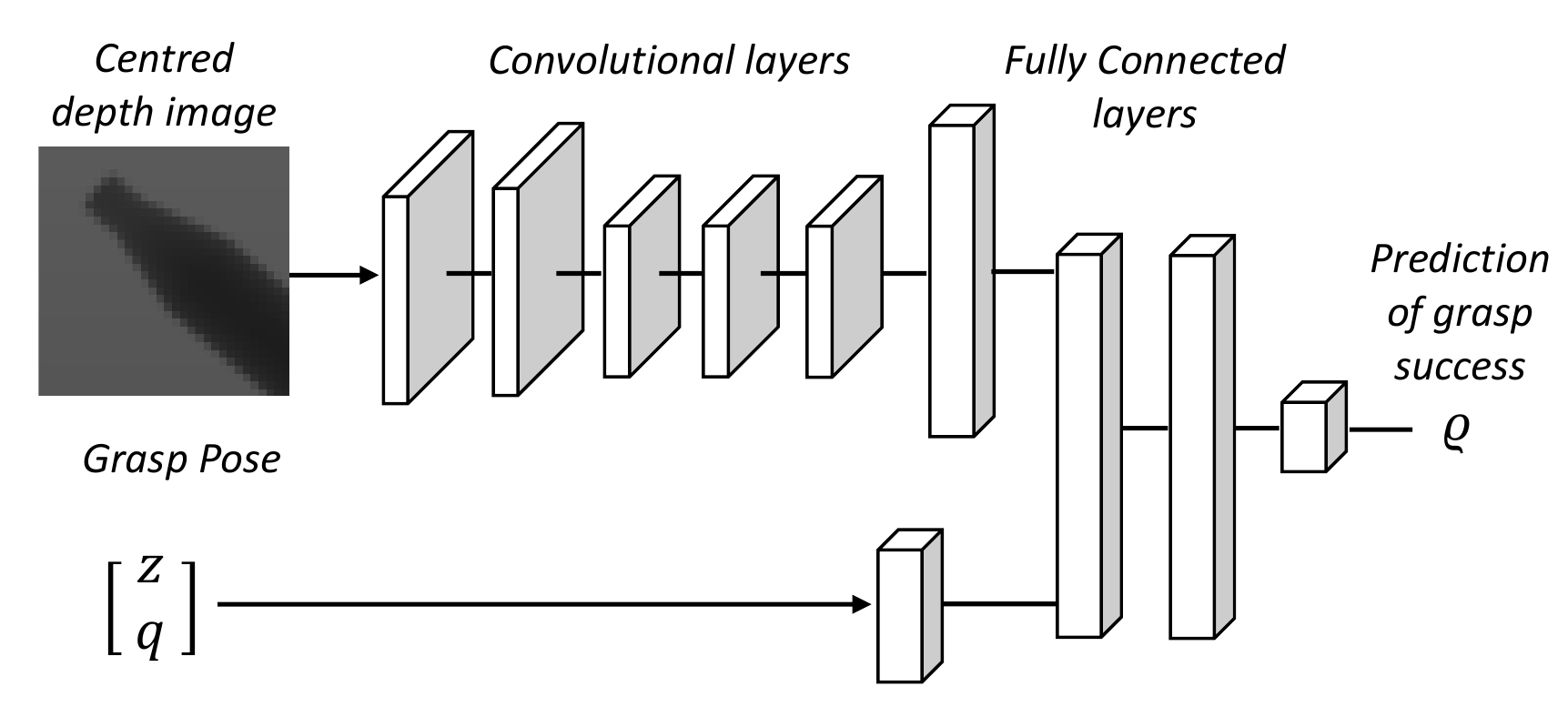}
      \caption{Architecture of \ours. The network consists of an image stream (upper), a pose stream (lower) and a merge stream (down-end) to predict grasp success. The input image is a $32 \times 32$ pixel depth image and the pose is given by the distance between the camera and the gripper TCP, $z$, and the orientation quaternion, $\quaternion \in R^4, \text{where } |q| = 1$. It outputs a prediction of grasp success $\prediction \in [0, 1]$.}
      \label{fig:mrgqcnn}
\end{figure}

In DexNet2.0~\cite{mahler2017dexnet2}, as well as in other image based grasp predictors~\cite{redmon2015, morrison2019, kumra2020, depierre2018jacquard, lenz2015}, the TCP of each grasp proposal is projected into the image to be represented in the {2-D} image plane, e.g. as an oriented rectangle. In these approaches, the grasp orientation is constrained to some arbitrary axis that is not explicitly specified as a network input. To represent grasp orientations that are not constrained to align with an arbitrary axis (such as the camera principal ray), requires an alternative grasp specification for the network. 

To represent the varying grasp orientations in \ours\, we use a grasp representation that defines the full 3-DOF grasp orientation in the form of a quaternion $\quaternion$. (While we use quaternions as orientation representations, also used in VGN~\cite{breyer2020volumetric}, this could be exchanged with other representations like Euler angles.) 

Using the approach of \gqcnn\ we specify the grasp image coordinates $(u, v)$ implicitly in the image. This is achieved by projecting the gripper TCP into the image $\mathcal{I}$, cropping the image around the grasp coordinates $(u, v)$ and subsequently resizing it to a $32 \times 32$ pixel image. The remaining variables of the {6-DOF} grasp, the distance of the TCP to the camera, $z$, and the grasp quaternion, $q$, are provided to \ours\ as an input in the pose stream, see Fig~\ref{fig:mrgqcnn}.

We keep the general network structure of \gqcnn, consisting of parallel image- and pose-streams with convolutional and fully connected layers. The two streams are combined into a single stream towards the end of the network. Apart from including the quaternion as an extra input, we add another fully-connected layer with 1024 nodes before the last layer. This extra layer adds representational capacity to account for the additional grasp information and increases network performance by $4\%$ as shown in section~\ref{sec:res_ablation}. \ours\ has a total of 6.5m and \gqcnn\ has a total of 5.4m trainable parameters. The full architecture of \ours\ can be seen in Fig.~\ref{fig:mrgqcnn}.


\begin{figure}[t]
      \centering
      \includegraphics[width=0.48\textwidth]{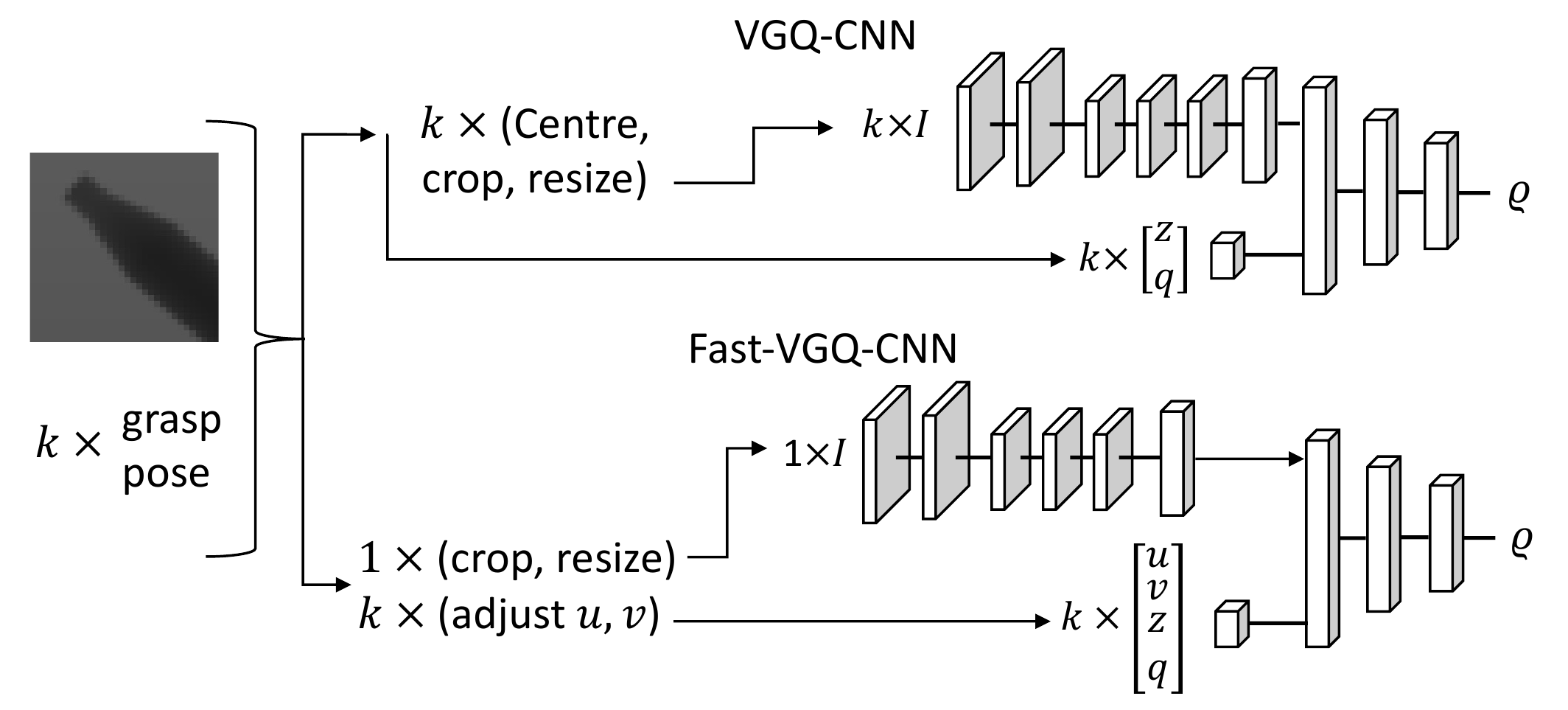}
      \caption{Comparison of pipelines for grasp quality evaluation with \ours\ and \fast. For \ours, $k$ images have to be generated from a single image and fed to the network with $k$ poses. For \fast, the image is processed once and fed into the image stream of the network. The output is then fed to the rest of the network together with $k$ poses.}
      \label{fig:speed}
\end{figure}

\section{\fast}\label{sec:speed}

While discriminative grasp proposal methods have several advantages, one of their main disadvantages is that there is typically a high latency for proposing grasps, as explained in section~\ref{sec:background}. For time-critical applications, we propose an alternative network architecture, \fast, that can decrease latency when evaluating multiple grasps on a given object. 

Instead of cropping the depth images such that the grasp centre is centred in the image, we decouple the grasp centre from the image cropping. To do this we make the $u$- and $v$-coordinates of the grasp centre within the image explicit inputs to the network. We apply random cropping~\cite{shorten2019augmentation} of the image around the grasp centre and specify the full 6-DOF pose input with $g \in (u, v, z, \quaternion)$ as an input to the pose stream of \fast.

Applying random cropping to the image ensures grasp- and object placement variety within the image during training, resulting in a network which can evaluate grasp proposals for varying grasp locations within the image. The coordinates of the grasp centre in the cropped and resized image $(u, v)$ are sampled uniformly. The range over which these coordinates are uniformly sampled, and therefore the maximum distance $\kappa = max(|u|, |v|)$ between the grasp centre and the image centre, influences the performance of \fast. This is shown in the ablation studies in section~\ref{sec:res_ablation}.

By introducing the new grasp representation, as well as the randomised grasp image coordinates $(u, v)$ during training, we can split the image stream with the high-dimensional convolutional layers from the rest of the network during inference. The image stream can be used as a shared encoder for multiple grasps, processing the image once and using the output for a batch of $k$ grasp poses. Utilising shared network layers to reduce run-time has been proposed and used for other network structures, e.g. for various object detection networks~\cite{zhao2019object}.

The new architecture also changes the requirements for image preprocessing starting from a single $300 \times 300$ pixel image and a variable number of grasp poses $k$, as indicated in Fig~\ref{fig:speed}. 
For each of the $k$ grasps in \ours, the image needs to be centred, cropped and resized, resulting in $k$ centred $32 \times 32$ pixel depth images and $k$ grasp poses $g \in (z, \quaternion)$. In contrast, for \fast, the original image needs to be cropped and resized just once. The grasp centre coordinates $(u, v)$ of each of the $k$ grasp poses are determined relative to the single image. Therefore the network input comprises a single $32 \times 32$ depth image and $k$ grasp poses $g \in (u, v, z, \quaternion)$. In both preprocessing pipelines, the image and pose values are normalised, excluding the quaternion.

For prediction, the resulting images and poses are fed to the network. For \ours, the $k$ depth images $\mathcal{I}$ and $k$ grasp poses go into a single, multi-stream network. For \fast, the single, depth image $\mathcal{I}$ goes into the shared encoder while the resulting tensor in combination with the $k$ grasp poses are fed to the remaining fully connected layers of the network.

\section{Experiments}\label{sec:results}

We train both \ours\ and \gqcnn\ for 150 million training iterations on \dset\ and DexNet2.0, respectively. One training epoch on a DexNet2.0-sized dataset is equivalent to approximately $6$ million training iterations. The training takes approximately $23$ hours to complete on one NVIDIA RTX 2060 GPU. We use a stochastic gradient decent optimiser with a momentum rate of $0.9$, a base learning rate of $0.001$ decaying every 4 million iterations by $0.95$, a $L2$-regularisation of $0.0005$ and sparse categorical cross-entropy loss.

In standard fashion, the depth images and $z$ values are normalised before being fed into the network. The values of the quaternion $q$ are not normalised, since they range within $-1,1$ naturally. We check performance on the validation split every 1 million iterations. For evaluating the resulting networks we use the balanced accuracy metric, which is calculated as the mean of true positive rate (TPR) and true negative rate (TNR). We motivate the use of the balanced accuracy metric due to the high imbalance of positive and negative grasps, with a network classifying all grasps as negative achieving an accuracy of $81\%$ due to that imbalance. We choose the best model according to the balanced accuracy metric of the validation results for all our evaluations.

We investigate the performance of \ours\ on a test split of \dset\ and compare performance between \gqcnn\ and \ours\ on a dedicated, DexNet2.0-like test set. In addition, we show the robustness of \ours\ to varying camera poses, proving that \ours\ does not need retraining when moving the camera within the range of the parameters specified in Table~\ref{tab:dset}. In a set of ablation studies, we show the effects of dataset size and the extra fully-connected layer on \ours, as well as the effects of the range of randomised grasp image coordinates $(u, v)$ on \fast. Finally, we demonstrate the speed up that can be achieved by using a shared image encoder in \fast.

\begin{table}[t]
\caption{Results}
\vspace{-0.3cm}
\label{tab:res}
\begin{center}
\begin{tabular}{c c|c c c }

Network   & Evaluation & TPR & TNR & Balanced\\
        & dataset    &     &     & accuracy\\
\hline
\gqcnn & DexNet2.0 & $70.0\%$ & $89.1\%$ & $79.5\%$\\

\ours  & \dset & $73.9\%$ & $90.2\%$ & $82.1\%$\\

\gqcnn & \ReducedDset & $63.6\%$ & $89.6\%$ & $76.7\%$\\

\ours & \ReducedDset & $64.6\%$ & $85.7\%$ & $75.2\%$\\

\end{tabular}
\end{center}
\end{table}

\subsection{Overall performance}\label{sec:res_overall}

For measuring the overall performance, we evaluate \ours\ on the test split of \dset\ (700K grasps), and \gqcnn\  on the test split of DexNet2.0 (700K grasps). In addition, to compare performance of \ours\ and \gqcnn, we evaluate them on a separate, new dataset with 300K grasps named Top Grasp Testset (\ReducedDset). The objects in \ReducedDset\ have not been seen by \ours\ or \gqcnn\ before, since they belong to the test split used in DexNet2.0 and \dset. \ReducedDset\ is rendered using the grasp and camera parameters of DexNet2.0 as detailed in Table~\ref{tab:dset}.

Due to the different input requirements in the networks, images are centred and rotated before being fed to \gqcnn\ and only centred for use in \ours. Note that the collision checking for \dset\ differs slightly from the approach used for DexNet2.0. In DexNet2.0, grasps are classified as collision free if any of four linear approaches within $\pm 10\degree$ around $\grasphinge$ are collision free. In \dset, grasps are classified as collision free only if a linear approach along the z-axis is collision free. For the purpose of comparison between \gqcnn\ and \ours, we exclude grasps from \ReducedDset\ which differ in outcome of the collision checking by these two methods, corresponding to $0.3\%$ of all generated grasps.

We calculate the true positive rate (TPR), true negative rate (TNR) and balanced accuracy. The results can be seen in Table \ref{tab:res}. Note that the balanced accuracy results for \gqcnn\ are lower than the accuracy reported in~\cite{mahler2017dexnet2}, since they used different training parameters and report accuracy rather than balanced accuracy.

\ours\ exhibits robust performance on \dset\ with a balanced accuracy of $82.1\%$, while being able to generalise to a wide range of camera poses and {6-DOF} grasp orientations. On the much restricted range of camera poses and grasp orientations in \ReducedDset, \ours\ achieves competitive performance with a balanced accuracy of $75.2\%$ compared to \gqcnn's $76.7\%$. Note that the performance of \fast\ depends on $\kappa$, as shown in section~\ref{sec:res_ablation}.

\subsection{Performance over the camera parameter space}\label{sec:param_res}

\begin{figure}[tb]
      \centering
      \includegraphics[width=0.5\textwidth]{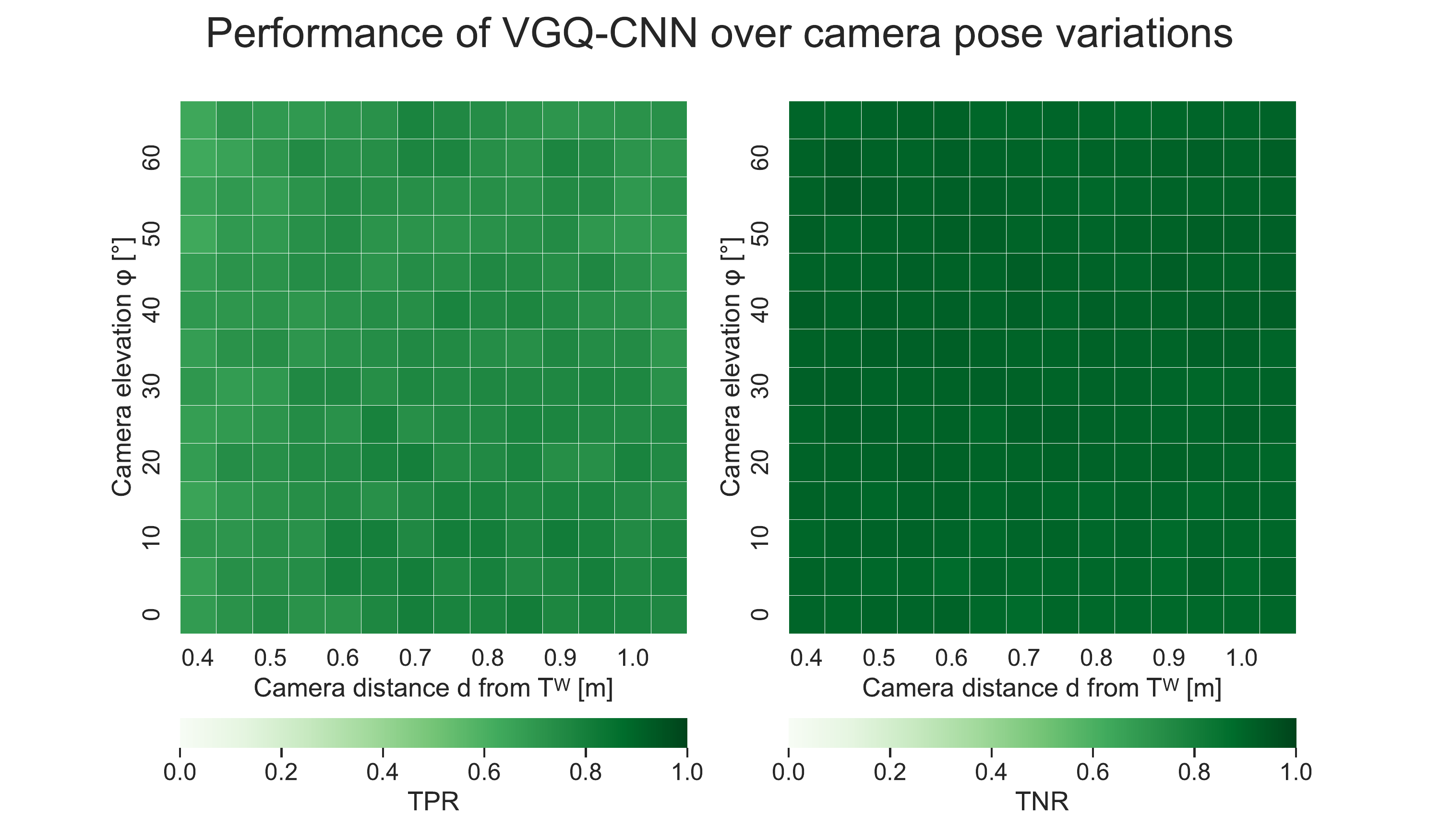}
      \caption{TPR (left) and TNR (right) for \ours, tested on an object-wise test split of \dset.}
      \label{fig:resgrid}
\end{figure}

In addition to \ours\ producing good results in \dset\ and being comparable to \gqcnn\ on \ReducedDset, it generalises well to varying camera positions. We report the TPR and TNR over the spherical coordinate variables of the camera, $\cameraradius$ and $\elev$, to show the applicability of \ours\ over varying camera poses in Fig.~\ref{fig:resgrid}. Over the full target range of camera poses with steps of $\Delta\cameraradius = 0.05 m$ and $\Delta\elev = 5\degree$, \ours\ attains $\mathit{TNR} = 90.2\% \pm 0.9\%$ and $\mathit{TPR} = 73.9\% \pm 3.5\%$. 

This robustness of \ours\ in terms of camera poses shows that it does not require retraining for new camera poses within its target range. Hence, \ours\ can be used for grasp quality prediction when moving camera poses and can even be used with mobile manipulators.

\subsection{Ablation studies}\label{sec:res_ablation}

\begin{figure}[tb]
      \centering
      \includegraphics[width=0.48\textwidth]{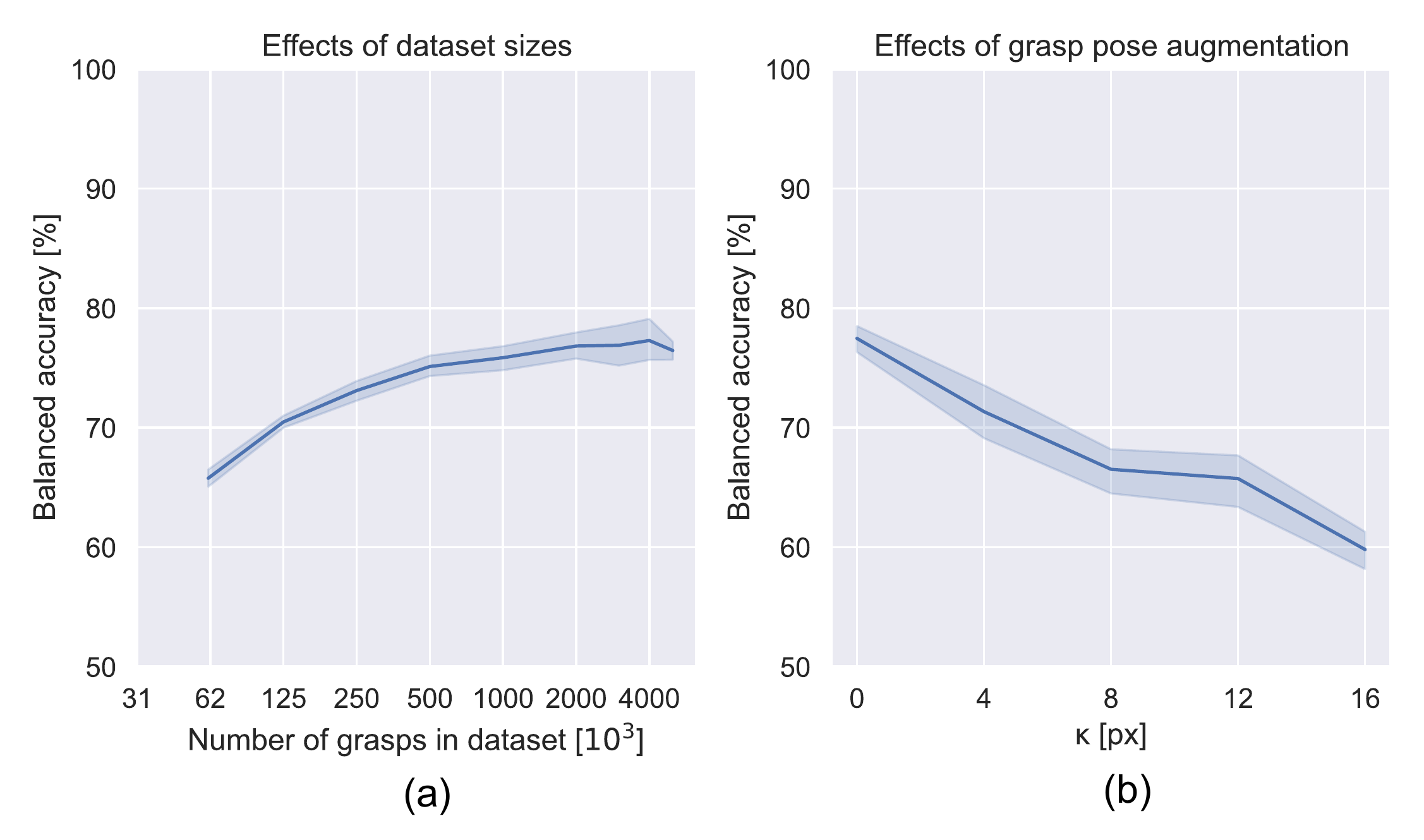}
      \caption{Influence of the dataset size (a) and maximum grasp distance $\kappa = max(|u|, |v|)$ to image centre (b) on network performance.}
      \label{fig:res_ablation}
\end{figure}

For further insight into the network performance, we conduct a set of ablation studies on the effect of the dataset size on \ours, the extra fully-connected layer in \ours\ and the randomised grasp image coordinates $(u, v)$ used for training our alternative network architecture, \fast. Note that all networks are trained from scratch for 10 million training iterations. 
Each network configuration was trained 8 times with the shading in Fig.~\ref{fig:res_ablation} corresponding to the $95\%$ confidence interval over the results. The datasets for the experiments of the dataset size influence were randomly undersampled from \dset. While the training and validation data differs for all networks in the dataset size experiments, they were tested on the same test split of \dset\ to ensure comparability. The base dataset for the experiments regarding the grasp image coordinates is \dset, with different $\kappa$ applied to the same train-validation-test split. 

Fig.~\ref{fig:res_ablation} (a) shows how the number of grasp training samples affects the trained network performance. The minimum number of training samples for reasonable network performance is relevant especially when evaluating different sampling strategies in section~\ref{sec:dset_sampling}, as some of these result in fewer training samples being available. When training the network with 1 million or fewer grasps, the network shows signs of overfitting and accuracy on the test set drops. Training \ours\ with 2 million grasps or more shows relatively stable results with a mean balanced accuracy of $77\%$. Note that a balanced accuracy of $50\%$ can be achieved by classifying all grasps as positive or negative.

Adding a second fully-connected layer with 1024 nodes before the last layer, as described in section~\ref{sec:architecture}, increases balanced accuracy from $72.5\% \pm 1.8\%$ to $76.7\% \pm 1.1\%$.

When using \fast, the grasp image coordinates are given as an explicit input to the network with $(u, v)$ and $\kappa = max(|u|, |v|)$ describes the maximum displacement of the grasp centre within the image. $\kappa = 0 px$ corresponds to the data used for \ours\ since each grasp is centred in its image. Fig.~\ref{fig:res_ablation} (b) exhibits a reduction in balanced accuracy as the displacement of the grasp centre in the image increases. For $\kappa = 0px$, \fast\ reaches a balanced accuracy of $77\%$ which drops to $60\%$ for grasps uniformly distributed over the full image with $\kappa = 16 px$.

\subsection{\fast}\label{sec:res_fast}

In section~\ref{sec:speed} we propose speeding up inference by using a shared encoder with our modified network architecture \fast. We measure the inference time of \ours\ and \fast\ on a machine with a 2.6 GHz Intel Core i7-10750 CPU and a NVIDIA GeForce RTX 2060 GPU. Tensorflow 1.15 is used for all experiments. We run the tests with differing numbers of grasps, $k \in \{32, 64, 96, 128\}$, as batch sizes and measure inference as the mean over 1000 runs. All image preprocessing steps are implemented with the pillow library in python. The resulting inference times for preprocessing and prediction can be seen in Table~\ref{tab:speed}.

\begin{table}[t]
\caption{Inference time [ms]}
\vspace{-0.3cm}
\label{tab:speed}
\begin{center}
\begin{tabular}{c | c c c c } 

& \multicolumn{4}{c}{Prediction} \\
$k$ & 32 & 64  & 96 & 128 \\
\hline
CPU \ours & $41$ & $72$ & $103$ & $138$\\

CPU \fast & $\mathbf{11}$ & $\mathbf{11}$ & $\mathbf{12}$ & $\mathbf{12}$ \\

GPU \ours & $8$ & $12$ & $17$ & $21$\\

GPU \fast & $8$ & $\mathbf{8}$ & $\mathbf{8}$ & $\mathbf{8}$\\

& \multicolumn{4}{c}{Preprocessing}\\

\ours & $7$ & $14$ & $21$ & $28$ \\

\fast & $\mathbf{0.3}$ & $\mathbf{0.5}$ & $\mathbf{0.6}$ & $\mathbf{0.8}$\\

\end{tabular}
\end{center}
\end{table}

The speed up of \fast\ relative to \ours\ increases as we increase the number of grasps predicted per batch, $k$. The \fast\ preprocessing pipeline is up to $35$ times faster than the preprocessing for \ours, taking just $0.8ms$ to preprocess 128 grasps. 

Of special interest when working with devices without a specialised GPU, \fast\ can predict 128 grasps within $12ms$, while \ours\ shows a significant increase for the prediction of more grasps taking $138ms$ for predicting 128 grasps on a CPU. In combination with an efficient grasp sampling strategy, \fast\ could enable real-time grasping.

\section{Discussion \& future work}\label{sec:conclusion}

In this work we present \ours, a network for predicting the quality of {6-DOF} grasps as observed from a wide range of camera poses. Removing limitations for grasp orientation and camera pose in previous methods~\cite{mahler2017dexnet2, levine2018, zhang2020randomforest, redmon2015, kumra2020, song2020graspdetection} and available grasp datasets~\cite{mahler2017dexnet2, depierre2018jacquard, lenz2015} allows for \ours\ to be used with {6-DOF} grasps and a wide range of camera poses without the need to retrain the network. This is especially useful when working with mobile manipulators, which constantly change the relationship between the camera and object.

\ours\ achieves a balanced accuracy of $82.1\%$ on a test split of \dset\ while being able to generalise to camera poses within $2.1m^3$ above the planar surface. Further, we make a non-sampled version of our dataset \dset\ available for public use to train alternative versions of \ours, e.g. focusing on special camera or grasp configurations. By using a shared encoder with our alternative network architecture \fast, we can predict 128 grasps within $12ms$ using a CPU compared to $138ms$ with \ours. Although our ablation studies in section~\ref{sec:res_ablation} show that there is a trade-off between network performance and the number of grasps that can be presented in a single image (which is related to  $\kappa$), this speed up could be of particular interest when working with edge devices without dedicated GPUs.

In common with other approaches described in the literature, our approach has some limitations. One of these is the position of the objects on a table below the camera. While this is suitable for objects placed on normal tables, it does not reflect situations where the camera is level with or below the object such as when a mobile manipulator should fetch an object from a shelf. The current network also excludes objects being placed very close to the edge where the table edge would be visible in the depth image.

As \ours\ is a grasp quality predictor, we aim to incorporate it with a {6-DOF} grasp sampling technique to generate a full grasp proposal pipeline in future work. We then plan to apply this full pipeline to a mobile manipulator, e.g. the {PAL} TIAGo robot~\cite{pal2016tiago}, and test performance on a grasping benchmark dataset such as EGAD~\cite{morrison2020egad}. Of special interest here is the comparison to point-cloud based {6-DOF} grasp proposal systems like GPD~\cite{pas2017pointclouds}, with differences in run-time and performance being crucial indicators of their usage for real-world applications.

\bibliographystyle{IEEEtran}
\bibliography{main}

\end{document}

%% file: nomenclature.tex
\newcommand{\elev}{\varphi} 
\newcommand{\polar}{\theta} 
\newcommand{\cameraradius}{d} 

\newcommand{\graspcameraangle}{\Psi}

\newcommand{\grasptableangle}{\beta}

\newcommand{\grasphinge}{\omega}
\newcommand{\graspquality}{\rho}
\newcommand{\quaternion}{q}
\newcommand{\prediction}{\varrho}